\documentclass[letterpaper]{article} 
\usepackage{aaai23}  
\usepackage{times}  
\usepackage{helvet}  
\usepackage{courier}  
\usepackage[hyphens]{url}  
\usepackage{graphicx} 
\usepackage{natbib}  
\usepackage{caption} 
\usepackage{algorithm}
\usepackage{algorithmic}

\usepackage{adjustbox}
\usepackage{multirow}
\usepackage{booktabs}
\usepackage{amsmath}
\usepackage{amsfonts}
\usepackage{siunitx}
\newcommand{\cls}{[\textsc{cls}]}
\newcommand{\sep}{[\textsc{sep}]}

\DeclareUnicodeCharacter{2212}{-}

\usepackage{newfloat}
\usepackage{listings}
\DeclareCaptionStyle{ruled}{labelfont=normalfont,labelsep=colon,strut=off} 
\floatstyle{ruled}
\newfloat{listing}{tb}{lst}{}
\floatname{listing}{Listing}
\title{RWEN-TTS: Relation-aware Word Encoding Network for Natural Text-to-Speech Synthesis}
\author{
    Shinhyeok Oh\equalcontrib,
    HyeongRae Noh\equalcontrib, 
    Yoonseok Hong,
    and Insoo Oh
}
\affiliations{
    Netmarble AI Center

}

\begin{document}

\maketitle

\begin{abstract}
With the advent of deep learning, a huge number of text-to-speech (TTS) models which produce human-like speech have emerged. Recently, by introducing syntactic and semantic information w.r.t the input text, various approaches have been proposed to enrich the naturalness and expressiveness of TTS models. Although these strategies showed impressive results, they still have some limitations in utilizing language information. First, most approaches only use graph networks to utilize syntactic and semantic information without considering linguistic features. Second, most previous works do not explicitly consider adjacent words when encoding syntactic and semantic information, even though it is obvious that adjacent words are usually meaningful when encoding the current word. To address these issues, we propose Relation-aware Word Encoding Network (RWEN), which effectively allows syntactic and semantic information based on two modules (i.e., Semantic-level Relation Encoding and Adjacent Word Relation Encoding). Experimental results show substantial improvements compared to previous works.

\end{abstract}
\section{Introduction}
Text-to-Speech (TTS), which aims at synthesizing natural-sounding speech from text, has extensive applications in various industries such as entertainment, education, and so on~\cite{DBLP:journals/corr/abs-2106-15561}. Recently, deep learning-based TTS models have drawn attention, showing unprecedented results.
Most existing works have adopted a two-stage generation scheme, which produces an intermediate speech representation (e.g., Mel-spectrogram) from the input text and then generates a raw waveform.
In this work, we focus on the model used in the first stage, called an acoustic model. Generally, the acoustic model is categorized into the autoregressive (AR) model and the non-autoregressive (NAR) model, according to the generation method. Early studies usually focused on the AR model~\cite{DBLP:conf/ssw/OordDZSVGKSK16, DBLP:conf/icml/Skerry-RyanBXWS18, DBLP:conf/icassp/ShenPWSJYCZWRSA18, 10.1609/aaai.v33i01.33016706}. However, they have a slow inference speed caused by sequential generation. Moreover, they are quite sensitive to the alignment resulting in low robustness (e.g., long pause, word repeating, and word skipping).
To overcome these limitations, many NAR models~\cite{lancucki2021fastpitch, ren2019fastspeech, DBLP:conf/iclr/0006H0QZZL21} have been proposed. Compared to AR models, they showed faster inference speed by generating speech in parallel and alleviated robustness issues. Nevertheless, their quality of expressiveness is unsatisfactory because they predict prosodic features that contain pitch, duration, and energy without introducing dependency between time steps~\cite{kharitonov-etal-2022-text}. Thus, various approaches to improve the quality of NAR-TTS have been proposed. 
\begin{figure}
\includegraphics[width=0.49\textwidth]{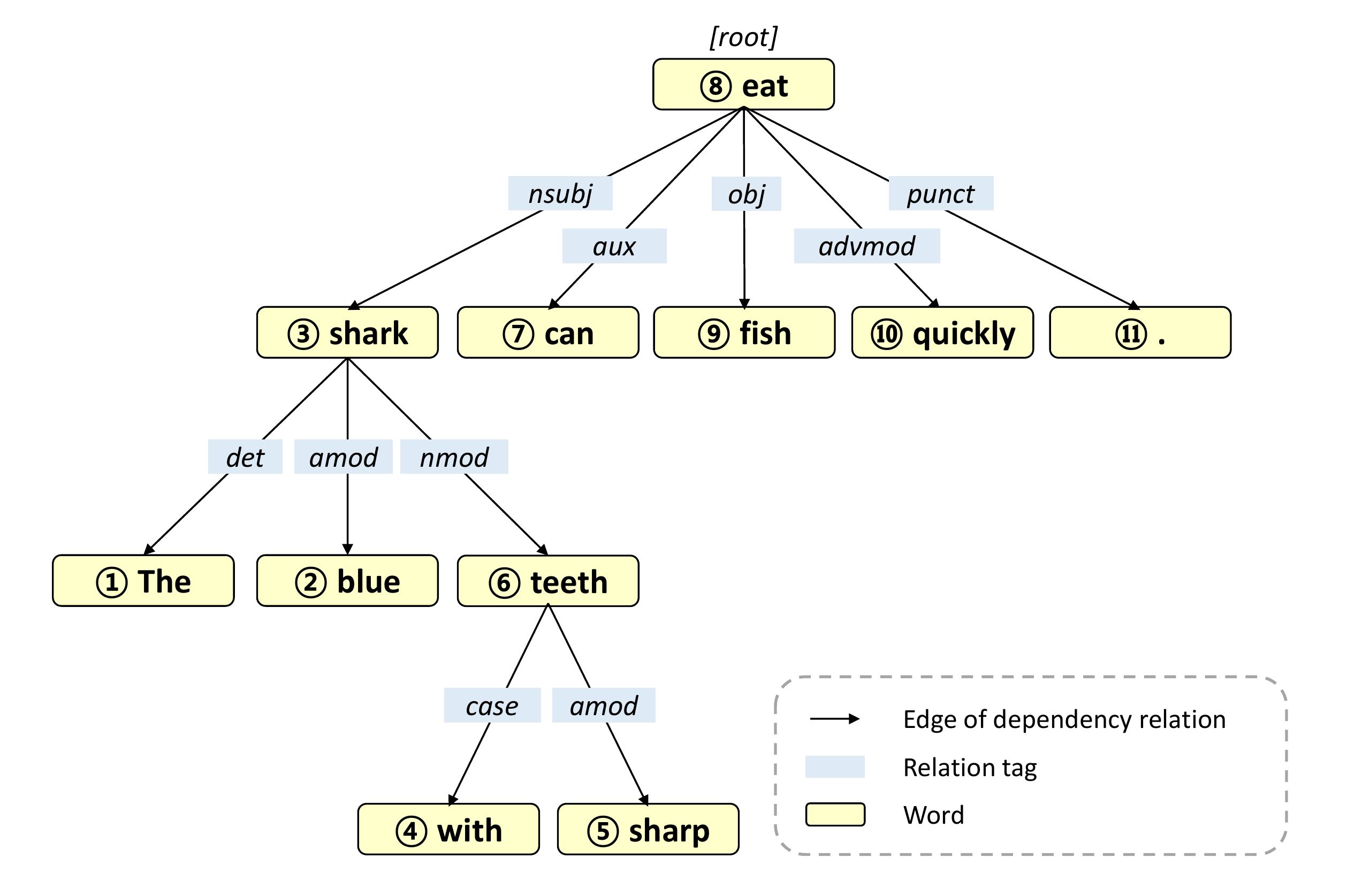}
\caption{An example of a dependency tree to illustrate for ``The blue shark with sharp teeth can eat fish quickly." The description of each element is described in the bottom right corner. For example, ``\textit{det}" is the relation tag between ``The" and ``shark".}
\label{fig:01_dependency_tree_example}
\end{figure}
\citet{min2021meta} successfully achieved expressive speech synthesis by introducing a reference encoder that models desired prosody because the same sentence can be uttered in diverse styles. 
\begin{figure*}[t]\centering
\includegraphics[width=\textwidth]{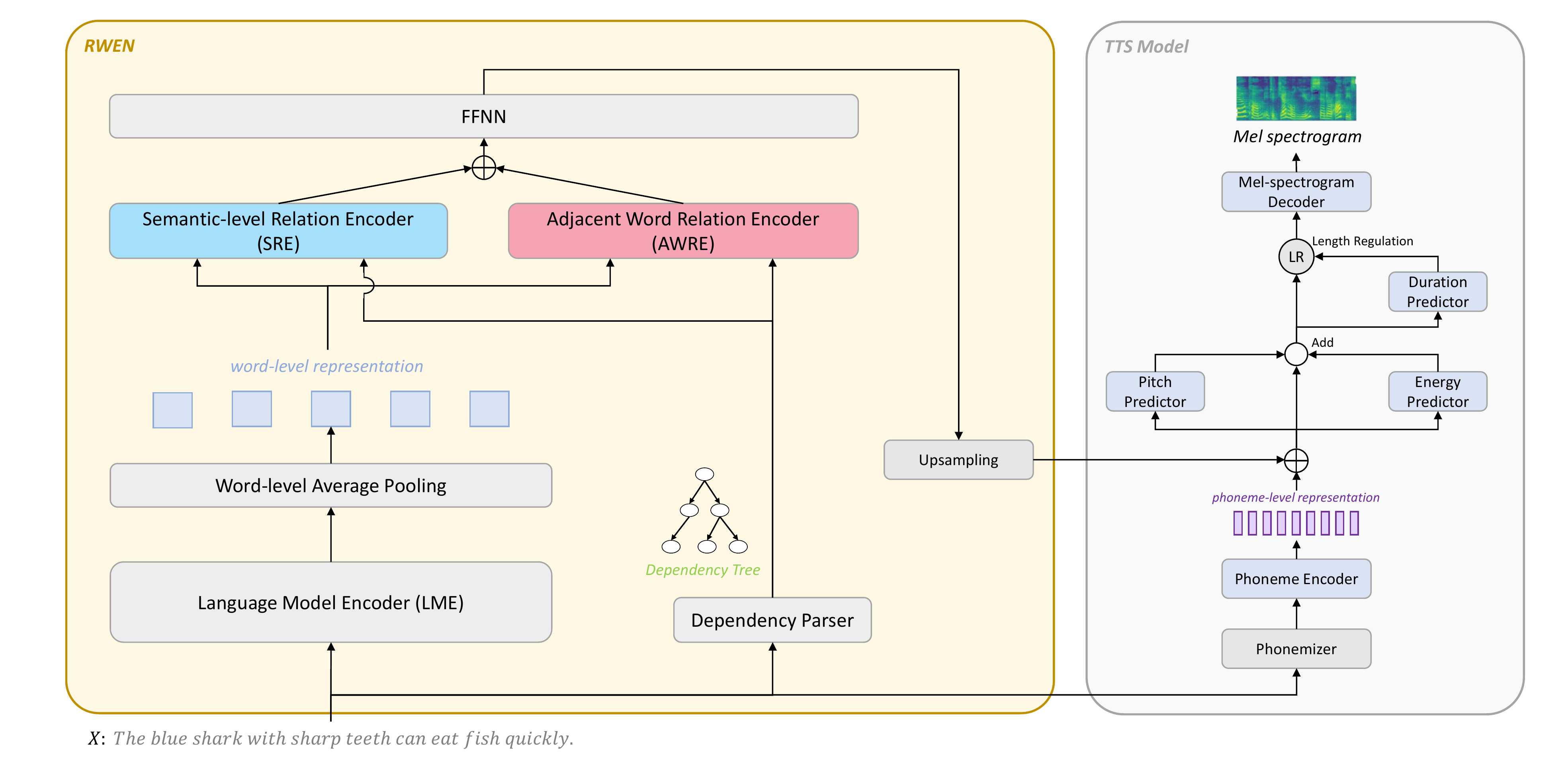}
\caption{Overall architecture of RWEN for TTS. \(\oplus\) denotes concatenation operator and \(\bigcirc\) denotes element-wise add operator. Length regulation refers to upsampling by repeating for each phonemic representation as much as the predicted duration.}
\label{fig:02_overall_architecture}
\end{figure*}
\citet{hwang2021tts, DBLP:conf/interspeech/SongYKSHOYKK22, lajszczak2022distribution} claimed that the performance of NAR-TTS is poor when the training data is insufficient, devising effective data augmentation methods.
\citet{kim2021conditional} combined powerful generative models (i.e., variational autoencoder, normalizing flow, and generative adversarial network) to improve expressiveness. They reported that proposed model close to human-level speech.
Meanwhile, \citet{DBLP:conf/interspeech/KenterSC20, 9413513, DBLP:conf/interspeech/JiaZSZW21, DBLP:conf/interspeech/ZhouSL0B0M22, DBLP:conf/interspeech/ZhangS0TYLWZQLZ22, tatanov2022mixer} boosted the expressiveness of speech by applying various methods proposed in the field of natural language processing (NLP) to the speech domain.
Especially, GraphSpeech~\cite{9413513} and Relational Gated Graph Network (RGGN)~\cite{DBLP:conf/interspeech/ZhouSL0B0M22} claimed the syntactic and semantic information of text affects the naturalness and expressiveness of speech. They improved the performance by utilizing graph networks focused on the representation based on dependency relations.

Despite the impressive results, we point out two crucial problems in applying syntactic and semantic information. First, most previous works utilizing dependency relations tend to assign graph networks to encode the neighbor nodes based on the dependency tree. For example, in Figure~\ref{fig:01_dependency_tree_example}, when encoding ``shark", RGGN utilizes weighted-sum to encode ``the", ``blue", and ``teeth", simultaneously. In RGGN, these neighbor words are explicitly considered, and others are implicitly considered. However, ``blue" and ``teeth" do not have a direct semantic correlation, except they share the same parent. We assume that encoding dimly correlated words simultaneously and explicitly can confuse the model. Second, previous works do not explicitly consider dependency relations on adjacent words. On the other hand, it is obvious the relations of adjacent words are usually meaningful because the TTS task deals with sequential data.


To address the aforementioned issues, we propose Relation-aware Word Encoding Network (RWEN) for TTS. RWEN, which consists of Semantic-level Relation Encoding (SRE) and Adjacent Word Relation Encoding (AWRE), focuses on effectively encoding dependency relations to improve naturalness and expressiveness. SRE encodes dependency relations based on the semantic level to substitute the inefficient graph networks mentioned above. AWRE explicitly encodes dependency relations based on adjacent words. We briefly summarize our main contributions as follows:
\begin{itemize}
\item We design two novel approaches, SRE and AWRE, to consider linguistic features and TTS characteristics.
\item We propose RWEN that contains SRE and AWRE, which can be easily incorporated into most recent TTS models.
\item Experimental results demonstrate that RWEN outperforms existing works, and we prove that SRE and AWRE are significantly effective through our ablation experiments.
\end{itemize}

\section{Proposed Method}
\subsection{RWEN: Relation-aware Word Encoding Network}
Figure~\ref{fig:02_overall_architecture} describes the overall architecture of RWEN. As mentioned before, to solve two crucial problems when assigning dependency relations, we propose a novel approach called RWEN that contains SRE and AWRE.

\subsubsection{Task Description}
Given a text \( X\), we aim to generate the natural and expressive speech.

\subsubsection{Dependency Parser}
We utilize a dependency parser to get the dependency tree. The tree has dependency relations between words, as shown in Figure~\ref{fig:01_dependency_tree_example}.
The dependency parser takes an input sequence represented as,
\begin{equation}
\label{eq:1}
\mathbf{X}^{W} = [{X}^{W}_{1}\,{X}^{W}_{2}\,...\,{X}^{W}_{n}],
\end{equation}
where \( X^{W}\) represents the list of tokens divided on the basis of words and \( n\) denotes the number of words. And the output is a tree like the one described in Figure~\ref{fig:01_dependency_tree_example}. 
The output tree consists of heads and relation tags represented as, \(head = [head_{1}, head_{2}, ..., head_{n}]\) and \(rel = [rel_{1}, rel_{2}, ..., rel_{n}]\). To utilize dependency relations, we define the Relation Tag Embedding (RTE), which makes embeddings for each relation tag,
\begin{equation}
\begin{split}
\label{eq:2}
E^{T}_{i} &= \Phi_{RTE}(rel_{i})\\
E^{T} &= [E^{T}_{1}, E^{T}_{2}, ..., E^{T}_{i}, ..., E^{T}_{n}],
\end{split}
\end{equation}
where \( i\) is an index in the range of \( n\) and \(\Phi_{RTE}\) denotes the embedding look-up table for RTE. \(E^{T} \in \mathbb{R}^{d_{{E}^{T}}\times n}\) is embedded representations for one sentence, where \(d_{{E}^{T}}\) represents a dimension of \(\Phi_{RTE}\). Finally, ${E}^{T}$ is fed into SRE and AWRE.

\subsubsection{Language Model Encoder}
Following recent works, we utilize pre-trained language models, such as BERT~\cite{devlin-etal-2019-bert} and ELECTRA~\cite{DBLP:conf/iclr/ClarkLLM20}, as Language Model Encoder (LME) described in Figure~\ref{fig:02_overall_architecture} to construct text representation.
The input sequence for LME is represented as,
\begin{equation}
\label{eq:3}
\mathbf{X}^{S} = [\cls\,{X}^{S}_{1}\,X^{S}_{2}\,...\,X^{S}_{m}\,\sep],
\end{equation}
where \( X^{S}\) represents the list of tokens divided on the basis of subwords and \( m\) denotes the length of tokens for the input sentence tokenized by the pre-trained language model tokenizer. ${X}^{S}$ is fed into the pre-trained language model to obtain the output text representation, \(H^{S} = [H^{S}_{\cls}, H^{S}_{1}, H^{S}_{2}, ..., H^{S}_{m}, H^{S}_{\sep}] \in \mathbb{R}^{d_H\times (m+2)}\), where \(d_H\) represents a dimension of the pre-trained language model.

\subsubsection{Word-level Average Pooling}
While the dependency relations are divided based on words, \( H^{S}\) are split based on subwords. We need the proper way to align the dependency relations with \( H^{S}\) because we use them simultaneously. Therefore, we utilize Word-level Average Pooling to align between them, similar to Subword-to-Word Mapping by \citet{DBLP:conf/interspeech/ZhouSL0B0M22}. We use average pooling (\(AP\)) based on word level represented as,
\begin{equation}
\begin{split}
\label{eq:4}
H^{W}_{i} &= AP([{H}^{S}_{j}, {H}^{S}_{j+1}, ..., {H}^{S}_{z}])\\
H^{W} &= [H^{W}_{\cls}, H^{W}_{1}, ..., H^{W}_{i}, ..., H^{W}_{n}, H^{W}_{\sep}],
\end{split}
\end{equation}
where \( j\) and \( z\) are the start and end index on subword-level based on \({X}^{W}_{i}\), respectively. For example, if the word ``quickly" is tokenized as ``quick" and ``ly", \({H}^{W}_{i}\) can be represented as \({H}^{W}_{i}= AP([{H}^{S}_{quick}, {H}^{S}_{ly}])\), where \({H}^{S}_{quick}\) and \({H}^{S}_{ly}\) denote output text representations for indexes of ``quick" and ``ly" in \(H^{S}\). \(H^{W} \in \mathbb{R}^{d_H\times (n+2)}\) represents the word-level representation described in Figure~\ref{fig:02_overall_architecture}. \(H^{W}_{\cls}\) and \(H^{W}_{\sep}\) in \(H^{W}\) are equal to \(H^{S}_{\cls}\) and \(H^{S}_{\sep}\), respectively. Finally, \(H^{W}\) is fed into SRE and AWRE.

\begin{figure}
\includegraphics[width=0.49\textwidth]{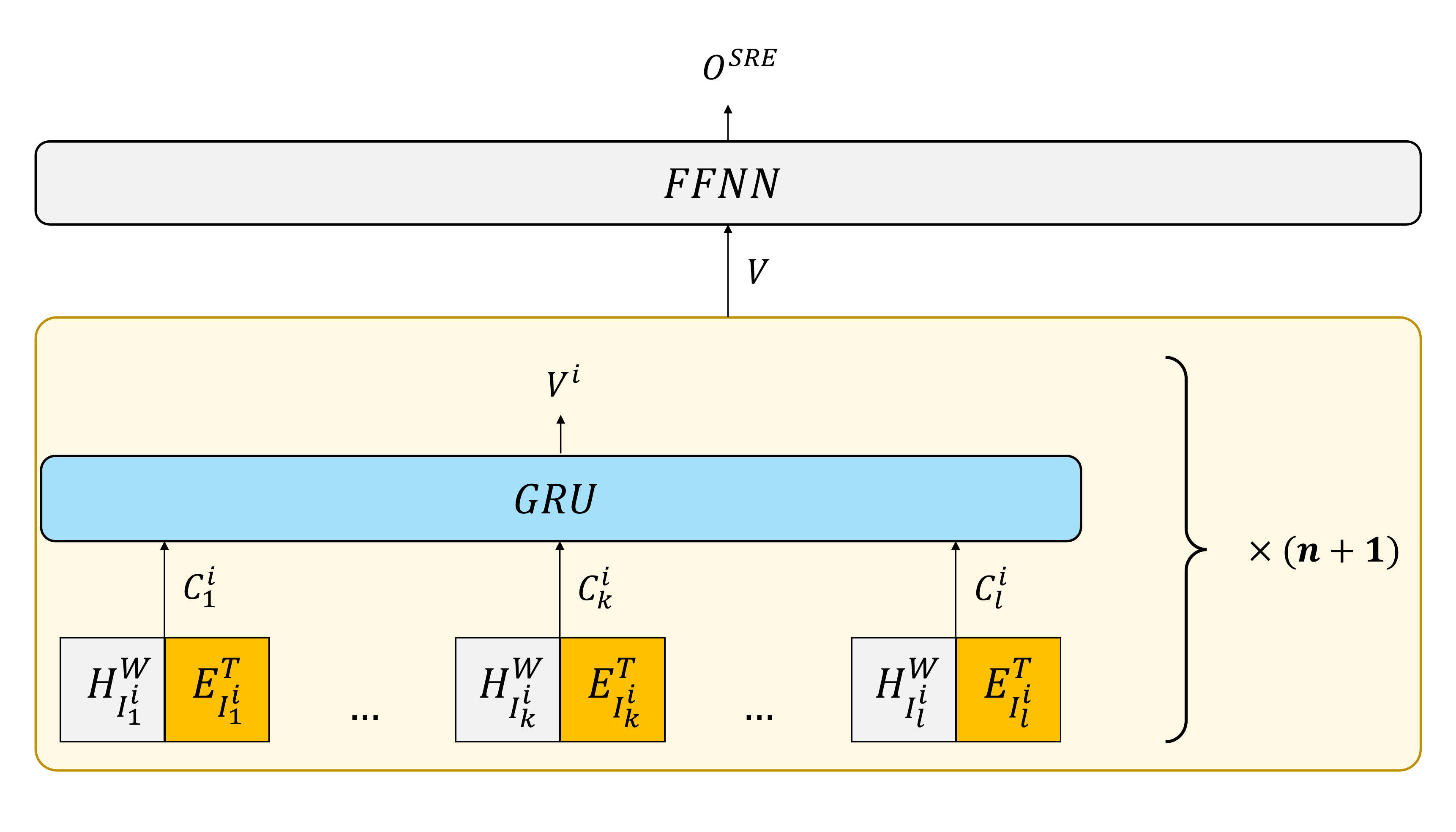}
\caption{The architecture of SRE. Each GRU encodes vectors concatenated with the word-level representation and RTE.}
\label{fig:03_sre_architecture}
\end{figure}
\subsubsection{Semantic-level Relation Encoding} 
Previous work~\cite{DBLP:conf/interspeech/ZhouSL0B0M22} proposes RGGN utilizing the graph networks based on the dependency tree. Since the feature of the dependency tree is related to the prosody of speech~\cite{DBLP:conf/interspeech/KohnBD18, 9413513, DBLP:conf/interspeech/ZhouSL0B0M22} and graph networks are suitable for encoding tree structures, they report improved results compared to baseline. However, they only use graph networks to encode neighbor words in the tree and don't seem to consider linguistic features. This method can encode the tree structure, but it is inefficient because linguistic features are not considered.
Therefore, we propose SRE to effectively encode phrases with contextual meaning. We assume that the phrase from each word to the root can be defined as phrases with their contextual meaning because they are sequentially connected in a dependency tree. SRE is described in Figure~\ref{fig:03_sre_architecture} and as follows.

SRE aims to encode phrases from each word to the root. Each word has a phrase with contextual meaning, and the indexes from the word to the root node are represented as,
\begin{equation}
\label{eq:5}
I^{i} = [I^{i}_{1}, ..., I^{i}_{l}], \\
\end{equation}
where \( l\) is the length from the current node \(i\) to the root node. \(I^{i}\) denotes word indexes in the phrase starting with the index \(i\) of each word. For example in Figure~\ref{fig:01_dependency_tree_example}, if the value of \( i\) is 2, \(I^2\) is represented as, \(I^{2} = \{2, 3, 8\}\). To expand to the sentence, the indexes can be represented as,
\begin{equation}
\label{eq:6}
I = [I^{0}, I^{1}, I^{2}, ..., I^{n}, I^{n+1}],
\end{equation}
where \(I^{0}\) is the index of \(H^{W}_{\cls}\) and \(I^{n+1}\) is the index of \(H^{W}_{\sep}\). Then, we utilize the word-level representation and RTE as follows:
\begin{equation}
\begin{split}
\label{eq:7}
C^{i}_{k} &= H^W_{I^{i}_{k}} \oplus E^{T}_{I^{i}_{k}}\\
C^{i} &= [C^{i}_{1}, C^{i}_{2}, ..., C^{i}_{k}, ..., C^{i}_{l}],
\end{split}
\end{equation}
where \(k\) is an index of \(I^{i}\), and \(\oplus\) denotes the concatenation operator. \(C^{i}\) represents a vector with the contextual meaning and dependency relations. \(C^{i}\) is fed into Gated Recurrent Units (GRU)~\cite{69e088c8129341ac89810907fe6b1bfe}, and the output is represented as,
\begin{equation}
\begin{split}
\label{eq:8}
V^{i} &= GRU(C^{i}) \\
V &= [V^{0}, V^{1}, ..., V^{n+1}],
\end{split}
\end{equation}
where \(V^{i}\) is the last hidden state of GRU. And, \(V\) is fed into single-layer feed-forward neural network (FFNN) as,
\begin{equation}
\label{eq:9}
O^{SRE} = \omega_{1} V + b_{1},
\end{equation}
where \(\omega_{1} \in \mathbb{R}^{d_H\times (d_H + d_{E^{T}})}\) and \(b_{1}\) are trainable parameters.

\begin{figure}
\includegraphics[width=0.49\textwidth]{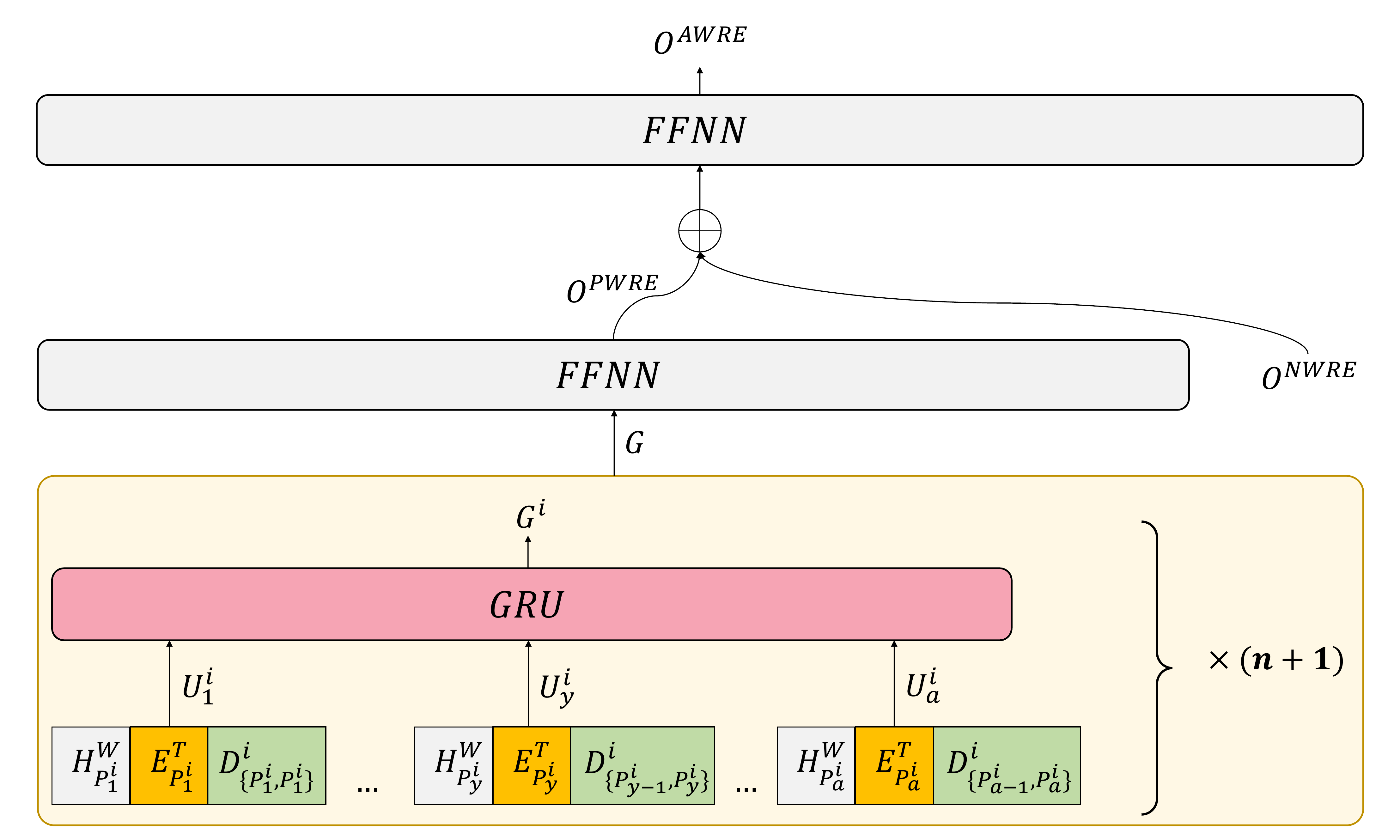}
\caption{The architecture of AWRE. Each GRU encodes vectors concatenated with the word-level representation, RTE, and DE.}
\label{fig:04_awre_architecture}
\end{figure}
\subsubsection{Adjacent Word Relation Encoding} 
To improve the results of TTS, we assign TTS characteristics as well as linguistic features. In particular, we focus that both input and output of TTS are sequential data, which are affected by surrounding words. To consider this, we propose AWRE, which encodes dependency relations between surrounding words and the current word based on the dependency tree. AWRE consists of two modules: Previous Word Relation Encoding (PWRE) and Next Word Relation Encoding (NWRE). PWRE encodes the dependency relations from the current word to the previous word, and NWRE encodes the dependency relations from the current word to the next word. First, PWRE is described in Figure~\ref{fig:04_awre_architecture} and as follows.

PWRE aims to encode dependency relations from the current word to the previous word. We construct the shortest path from the current word to the previous word in the dependency tree. Indexes in the shortest path are represented as,
\begin{equation}
\label{eq:10}
P^{i} = [P^{i}_{1}, ..., P^{i}_{y} , ..., P^{i}_{a}] \\
\end{equation}
where \(a\) is the length of the shortest path from the current node to the previous node. \(P^{i}\) denotes indexes starting from the current node \(i\) and ending with the previous node \(i-1\). For example in Figure~\ref{fig:01_dependency_tree_example}, if the value of \( i\) is 2, \(P^2\) is represented as, \(P^{2} = \{2, 3, 1\}\). To expand to the sentence, the indexes can be represented as,
\begin{equation}
\label{eq:11}
P = [P^{0}, ..., P^{i}, ..., P^{n}, P^{n+1}],
\end{equation}
where \(P^{0}\) is the index of \(H^{W}_{\cls}\) and \(P^{n+1}\) is the index of \(H^{W}_{\sep}\).

Additionally, we utilize directions between connected nodes based on the dependency tree represented as,
\begin{equation}
\label{eq:de}
Q_{\{P^{i}_{y-1},P^{i}_{y}\}} \in [self, parent, child],
\end{equation}
where \(Q_{\{P^{i}_{y-1},P^{i}_{y}\}}\) denotes the direction between \(P^{i}_{y-1}\) and \(P^{i}_{y}\). If the direction from \(P^{i}_{y-1}\) to \(P^{i}_{y}\) is the parent, \(Q_{\{P^{i}_{y-1},P^{i}_{y}\}}\) denotes \(parent\). Likewise, if the direction is the child, \(Q_{\{P^{i}_{y-1},P^{i}_{y}\}}\) denotes \(child\). When it needs to encode itself (i.e., first element), we use \(self\) to utilize the direction.
Then, we define Direction Embedding (DE) represented as,
\begin{equation}
\begin{split}
\label{eq:de2}
D^{i} = &\{\Phi_{DE}(self), ..., \Phi_{DE}(Q_{\{P^{i}_{y-1},P^{i}_{y}\}})\\&, ..., \Phi_{DE}(Q_{\{P^{i}_{a-1},P^{i}_{a}\}}))\},
\end{split}
\end{equation}
where \(\Phi_{DE} \in \mathbb{R}^{d_{\Phi_{DE}}\times 3}\) denotes the embedding look-up table for DE. \(\mathbb{R}^{d_{\Phi_{DE}}}\) represents a dimension of \(\Phi_{DE}\). DE is only used in AWRE. In SRE, we do not consider to use DE because it is encoded only in one direction based on the tree.

We utilize the word-level representation, RTE and DE as follows:
\begin{equation}
\begin{split}
\label{eq:12}
U^{i}_{y} &= H^W_{P^{i}_{y}} \oplus E^{T}_{P^{i}_{y}} \oplus D^{i}_{\{P^{i}_{y-1},P^{i}_{y}\}}\\
U^{i} &= [U^{i}_{1}, U^{i}_{2}, ..., U^{i}_{y}, ..., U^{i}_{a}],
\end{split}
\end{equation}
where \(U^{i}\) represents a vector with the contextual meaning, dependency relations, and the embedding of directions. \(U^{i}\) is fed into GRU, and the output is represented as,
\begin{equation}
\begin{split}
\label{eq:13}
G^{i} &= GRU(U^{i}) \\
G &= [G^{0}, G^{1}, ..., G^{n+1}],
\end{split}
\end{equation}
where \(G^{i}\) is the last hidden state of GRU. And, \(G\) is fed into FFNN as,
\begin{equation}
\label{eq:14}
O^{PWRE} = \omega_{2} G + b_{2},
\end{equation}
where \(\omega_{2} \in \mathbb{R}^{d_H\times (d_H + d_{E^{T}} + d_{\Phi_{DE}})}\) and \(b_{2}\) are trainable parameters.

NWRE is encoded similarly to PWRE, which has Equations~\ref{eq:10},~\ref{eq:11},~\ref{eq:de},~\ref{eq:12},~\ref{eq:13}, and~\ref{eq:14}. NWRE can be described by replacing the previous node only with the next node in the PWRE description. Thus, note that the final output of NWRE is represented \(O^{NWRE}\).
Then, \(O^{PWRE}\) and \(O^{NWRE}\) are concatenated as,
\begin{equation}
\label{eq:15}
O^{P\&N} = O^{PWRE} \oplus O^{NWRE},
\end{equation}
and fed into FFNN as,
\begin{equation}
\label{eq:16}
O^{AWRE} = \omega_{3} O^{P\&N} + b_{3},
\end{equation}
where \(\omega_{3} \in \mathbb{R}^{d_H \times (d_H + d_H)}\) and \(b_{3}\) are trainable parameters.

\subsubsection{Upsampling} 
\(O^{SRE}\) and \(O^{AWRE}\) are concatenated and fed into FFNN. Then, the output is represented by the word-level. We should concatenate with the output and phoneme-level representation, as shown in Figure~\ref{fig:02_overall_architecture}. However, phoneme-level representation is represented by the phoneme-level so that we can't directly concatenate. Thus, an upsampling method is required to concatenate with them. We duplicate the word-level segmented output representation by the number of phoneme sequences corresponding to each word and concatenate it with the phoneme-level representation.

\subsection{TTS Model}
To prove the effectiveness of our method, we adapt FastPitch~\cite{lancucki2021fastpitch} equipped with Unsupervised Alignment Learning framework (UAL)~\cite{badlani2022one} as the TTS model, which is one of the representative NAR-TTS models. More specifically, as shown in Figure~\ref{fig:02_overall_architecture}, it consists of five modules: Phoneme Encoder, Mel-spectrogram Decoder, Pitch Predictor, Energy Predictor, and Duration Predictor. Phoneme Encoder produces the phoneme-level representation from the phonemic text. Then, Pitch Predictor and Energy Predictor take the phoneme-level representation concatenated with the output representation of RWEN, constructing the pitch and energy information. With the help of UAL, Duration Predictor can be learned to predict the duration of each phoneme, which is used to perform upsampling from phoneme-level representation to frame-level one. Note that the prosody of synthesized speech can be controlled by adjusting the predicted pitch and duration during the inference stage. Finally, Mel-spectrogram Decoder generates the output Mel-spectrogram from the frame-level representation.

\section{Experiments}
\subsection{Experimental Setup}
\subsubsection{Datasets}
We train and evaluate RWEN on LJSpeech~\cite{ljspeech17}, a single speaker corpus recorded by a female English speaker. It consists of 13,100 short audio clips with a total length of 24 hours, being randomly split into 12,500, 100, and 500 samples to comprise the training, validation, and test datasets as in \citet{kim2021conditional}. Additionally, recent works~\cite{kim2021conditional, https://doi.org/10.48550/arxiv.2205.04421} have already achieved human-level performance on the benchmark datasets (e.g., LJSpeech, VCTK~\cite{vctk19}, etc.). Therefore, we evaluate RWEN on other type of datasets used in the field of NLP in order to derive meaningful comparison results.
To cover multiple domains, we evaluate RWEN on the following datasets:
\begin{itemize}
\item \textbf{CNN/Daily Mail}~\cite{nallapati-etal-2016-abstractive} contains articles from CNN and DailyMail newspapers.
\item \textbf{Children's Book Test (CBT)}~\cite{DBLP:journals/corr/HillBCW15} contains sentences built from books for children that are freely available.
\item \textbf{OpenBookQA}~\cite{mihaylov-etal-2018-suit} contains a small book of core elementary-level science facts.
\item \textbf{SQuAD 2.0}~\cite{rajpurkar-etal-2018-know} contains sentences on a set of Wikipedia articles.
\end{itemize}

\subsubsection{Subjective Evaluation}
We conducted the crowd-sourced listening test for Mean Opinion Score (MOS) and Comparative Mean Opinion Score (CMOS) on Amazon Mechanical Turk \footnote{\url{https://www.mturk.com/}}. We used at least 50 sentences randomly sampled from each dataset for all evaluations, and at least 15 listeners participated. To maintain evaluation quality, master workers certificated in Amazon Mechanical Turk only participated, and all submissions of workers who did not pass occasional hearing tests were rejected. For MOS, we evaluated naturalness and expressiveness on a 5-point scale from 1 to 5. For CMOS, we evaluated which one is more natural and more expressive on a 7-point scale from -3 to 3. Also, CMOS was measured between the baseline and a comparative model. Therefore, it is only possible to compare models between the baseline and a comparative model.

\begin{table}[t]\centering
\begin{adjustbox}{width=0.45\textwidth}
\begin{tabular}{lcc}
\toprule
 & MOS (CI) \\
\midrule
VITS~\cite{kim2021conditional} & \(\num{3.95}\) (\(\pm0.06\)) \\
FastPitch~\cite{lancucki2021fastpitch} & \(\num{3.74}\) (\(\pm0.06\)) \\
FastPitch w/ UAL & \(\num{4.04}\) (\(\pm0.06\)) \\
\bottomrule
\end{tabular}

\end{adjustbox}
\caption{Evaluation results for existing TTS models on the CNN/Daily Mail dataset. We measured with Mean Opinion Score (MOS) and 95\% confidence intervals (CI).}
\label{table:baseline_comparsion}
\end{table}

\begin{table*}[t]

\centering
\resizebox{0.95\textwidth}{!}{
\begin{tabular}{|l|c|c|c|c|c|}
\hline
\multirow{3}{*}{} & \multicolumn{5}{|c|}{MOS (CI)} \\
\cline{2-6}
&\multirow{2}{*}{LJSpeech}    & News               & \multicolumn{2}{|c|}{Book}     & Wiki     \\
\cline{3-6}
& & CNN/Daily Mail & CBT & OpenBookQA & SQuAD 2.0  \\ 
\hline
Ground Truth                      & 4.25 (\(\pm\)  0.06)   & -   & -   & -   & - \\
\hline
VITS                              & 4.04 (\(\pm\)  0.06)   & 4.01 (\(\pm\)  0.06)   & 3.94 (\(\pm\)  0.05)   & 3.98 (\(\pm\)  0.06)   & 4.03  (\(\pm\)  0.06) \\
FastPitch w/ UAL           & 4.16 (\(\pm\)  0.06)   & 4.05 (\(\pm\)  0.06)   & 4.06 (\(\pm\)  0.05)   & 3.91 (\(\pm\)  0.07)   & 4.10 (\(\pm\)  0.06) \\
RGGN-BERT    & 4.15 (\(\pm\)  0.06)   & 4.00 (\(\pm\)  0.06)   & 4.07 (\(\pm\)  0.05)   & 3.95 (\(\pm\)  0.07)   & 4.12 (\(\pm\)  0.06) \\
\hline
RWEN-BERT   & \textbf{4.19 (\(\pm\)  0.06)}   & \textbf{4.15 (\(\pm\)  0.06)}   & \textbf{4.15 (\(\pm\)  0.05)}   & \textbf{4.00 (\(\pm\)  0.06)}   & \textbf{4.18 (\(\pm\)  0.06)} \\
\hline

\end{tabular}}
\caption{Evaluation results of MOS with 95\% CI. The best scores except Ground Truth are in bold. `-' denotes the dataset doesn't have voices of Ground Truth or can't be evaluated because speakers are different between the training dataset and evaluation dataset. MOS was measured simultaneously within each column so that it is possible to compare models within the same column.
} 
\label{table:main_results}
\end{table*}
\subsubsection{Baselines}
For experiments, we compare our model with followings:
\begin{itemize}
\item \textbf{VITS}~\cite{kim2021conditional}
is a fully end-to-end TTS model that produces human-like sounding audio on the waveform domain by leveraging variational autoencoder (VAE)~\cite{DBLP:journals/corr/KingmaW13} with normalizing flows and adversarial training. To solve the one-to-many problem that one text can be spoken in various styles, they also introduced a flow-based stochastic duration predictor, demonstrating significant effectiveness. In this work, we used the official pre-trained model for fair comparisons~\footnote{\url{https://github.com/jaywalnut310/vits}}.
\item \textbf{FastPitch}~\cite{lancucki2021fastpitch}
is an acoustic model which generates a Mel-spectrogram from given text. It can control the pitch and duration of the synthesized speech by adjusting the outputs of pitch and duration predictors.
In this work, we used the official checkpoint~\footnote{\url{https://github.com/NVIDIA/DeepLearningExamples/blob/8d8c524df634e4dfa0cfbf77a904ce2ede85e2ec/PyTorch/SpeechSynthesis/FastPitch/scripts/download_fastpitch.sh}} for fair comparisons. And, since it generates a Mel-spectrogram from text, we need a model called vocoder that converts a Mel-spectrogram into a raw waveform. To this end, we use the official HifiGAN \cite{kong2020hifi} codes \footnote{\url{https://github.com/jik876/hifi-gan}} and a checkpoint pre-trained on the LJSpeech dataset and finetuned as the output of Tacotron 2~\cite{DBLP:conf/icassp/ShenPWSJYCZWRSA18}.
\item \textbf{FastPitch w/ UAL}  is a TTS model that contains FastPitch and Unsupervised Alignment Learning framework. We constructed by referring to codes of the official FastPitch repository~\footnote{\url{https://github.com/NVIDIA/DeepLearningExamples/tree/master/PyTorch/SpeechSynthesis/FastPitch}}. In addition, we modified the source code so that it can be processed in phoneme-level sequences for fair comparisons with our proposed model and RGGN.
\item \textbf{RGGN-BERT} \cite{DBLP:conf/interspeech/ZhouSL0B0M22} proposed RGGN to improve the naturalness and expressiveness of synthesized speeches. They utilized dependency structure and pre-trained BERT~\cite{devlin-etal-2019-bert} embedding. For their experiments, they used Tacotron 2~\cite{DBLP:conf/icassp/ShenPWSJYCZWRSA18} as the TTS model. However, in this work, we implemented RGGN with FastPitch w/ UAL for fair comparisons.
\end{itemize}
\subsubsection{TTS system for RWEN}
To prove the effectiveness of our proposed method, we adopt FastPitch w/ UAL. Compared to the recent end-to-end TTS model (e.g., VITS), FastPitch w/ UAL has controllability in terms of pitch and duration, which can utilize various applications. Also, it is light and easy to conduct diverse experiments. Moreover, as shown in Table~\ref{table:baseline_comparsion}, FastPitch w/ UAL achieved the best performance by introducing UAL and phoneme-level encoding.

\subsection{Implementation Details}

We implemented our proposed model, called RWEN, using the PyTorch~\cite{adam2019pytorch} and Transformers\footnote{\url{https://github.com/huggingface/transformers}}~\cite{wolf2020transformers} library. We adopt BERT\(_{base}\) and ELECTRA\(_{base}\)~\cite{DBLP:conf/iclr/ClarkLLM20} as the LME for our experiments. In our experiments, RWEN-BERT and RWEN-ELECTRA denotes RWEN using BERT\(_{base}\) and ELECTRA\(_{base}\) as the LME, respectively. Following RGGN, we use Stanza~\cite{qi-etal-2020-stanza} to get the dependency tree. We use FastPitch w/ UAL as our TTS model to simplify the experiments. Specifically, Phoneme Encoder and Mel-spectrogram Decoder are composed of four Feed-Forward Transformer (FFT) blocks \cite{ren2019fastspeech} whose parameters are the same as described in \citet{DBLP:conf/iclr/0006H0QZZL21} except that the hidden size of the Mel-spectrogram decoder is 1024. Duration Predictor, Pitch Predictor, and Energy Predictor are the same architecture: two 1-D convolutions with kernel size 3 and 256/256 input/output channels, each followed by ReLU, LayerNorm, and Dropout with the probability of 0.1. To extract the target pitch from the speech, we use the pYIN~\cite{DBLP:conf/icassp/MauchD14} algorithm and perform normalization with the mean and standard deviation of the pitch for the whole training dataset. Also, the energy of speech is extracted by performing the L2 norm on the Mel-spectrogram. The last fully connected layer projects a 256-dimensional vector into a scalar. To produce raw waveform from the synthesized Mel-spectrogram, pre-trained HifiGAN \cite{kong2020hifi} is used as the vocoder. Besides, we utilized the phonemizer \cite{Bernard2021} since we used the phoneme sequence as the input. We use mixed precision training on 16 Tesla A100 GPUs for all the experiments. The batch size is set to 2 per GPU, and the model is trained up to 200k steps. More details and samples are in our repository\footnote{\url{https://github.com/shinhyeokoh/rwen}} and demonstration site\footnote{\url{https://shinhyeokoh.github.io/demo/rwen_tts/index.html}}.

\begin{table}[t]\centering
\begin{adjustbox}{width=0.45\textwidth}
\begin{tabular}{lcc}
\toprule
 & CMOS & Wilcoxon p-value\\
\midrule
RWEN-BERT & $\num{0}$ & -\\
\midrule
\ \ \ w/o\ AWRE  & $\num{-0.09}$ & 7.5e-5 \\
\ \ \ w/o\ SRE  & $\num{-0.11}$ & 2.6e-6 \\
\ \ \ w/o\ SRE \& AWRE  & $\num{-0.13}$ & 4.1e-7\\
\bottomrule
\end{tabular}

\end{adjustbox}

\caption{Ablation study on the CNN/Daily Mail dataset. We choose RWEN-BERT as the baseline. We measured with CMOS and Wilcoxon p-value obtained by Wilcoxon signed rank test~\cite{Wilcoxon1992}.}
\label{table:ablation}
\end{table}
\subsection{Overall Results}
Table~\ref{table:main_results} reports MOS results to evaluate on LJSpeech, CNN/Daily Mail, CBT, OpenBookQA, and SQuAD 2.0 datasets. We observe that RWEN-BERT shows slightly lower MOS than Ground Truth in the LJSpeech evaluation dataset, and RWEN-BERT outperforms the comparative models for all datasets. This suggests that the proposed approaches are effective for TTS. 
For all datasets except OpenBookQA, FastPitch w/ UAL shows higher MOS than VITS. This can be additional evidence for Table 1, indicating that the naturalness and expressiveness of FastPitch w/ UAL are similar to or better than VITS.
FastPitch w/ UAL and RGGN-BERT show similar performance. As we mentioned in the introduction section, it can be seen that RGGN reflects syntactic and semantic information inefficiently.
As a result, RWEN-BERT achieves gains over FastPitch w/ UAL we utilized as our TTS system by 0.03 ($4.16 \rightarrow 4.19$), 0.10 ($4.05 \rightarrow 4.15$), 0.09 ($4.06 \rightarrow 4.15$), 0.09 ($3.91 \rightarrow 4.00$), and 0.08 ($4.10 \rightarrow 4.18$) on the LJSpeech, CNN/Daily Mail, CBT, OpenBookQA, and SQuAD 2.0 datasets, respectively.

\subsection{Ablation Study}
To study the effects of AWRE and SRE, we conduct ablation experiments on CNN/Daily Mail dataset. As shown in Table~\ref{table:ablation}, we set the baseline that RWEN-BERT. Removing AWRE (i.e., only utilizing SRE) and SRE (i.e., only utilizing AWRE) brings \(−0.09\) and \(−0.11\) CMOS degradation, respectively. If we remove SRE and AWRE (i.e., utilizing FastPitch w/ UAL), it brings \(−0.13\) CMOS degradation. We can observe CMOS drop significantly in all ablation experiments. This suggests all of our proposed approaches are effective for the TTS model. Meanwhile, we can also observe that the SRE is more effective than the AWRE. As SRE encodes phrases with contextual meaning, it allows the model to more exploit the dependency relations at a sentence-level.

\begin{table}[t]\centering
\begin{adjustbox}{width=0.45\textwidth}
\begin{tabular}{lcc}
\toprule
 & CMOS & Wilcoxon p-value\\
\midrule
RWEN-BERT & $\num{0}$ & - \\
\midrule
RWEN-ELECTRA & $\num{+0.22}$ & 1.4e-13 \\
\bottomrule
\end{tabular}

\end{adjustbox}

\caption{CMOS results on the CNN/Daily Mail dataset. To study effectiveness according to change of PLM, we choose RWEN-BERT as the baseline. We measured with CMOS and Wilcoxon p-value.}
\label{table:effect_lm}
\end{table}
\subsection{Effects of Pre-trained Language Model}
To study effectiveness according to the change of the pre-trained language model, we conduct CMOS experiments between RWEN-BERT and RWEN-ELECTRA. RWEN-ELECTRA uses ELECTRA\(_{base}\) as LME in our architecture. \citet{DBLP:conf/iclr/ClarkLLM20} reports that ELECTRA-base outperforms BERT-base on the GLUE~\cite{wang-etal-2018-glue} widely used benchmark for natural language understanding. 
As shown in Table~\ref{table:effect_lm}, RWEN-ELECTRA significantly improves compared to RWEN-BERT.
This suggests that the quality of the pre-trained language model affects RWEN, and using the improved language model has a positive effect on our proposed method.

\section{Conclusion}
In this study, we pointed out crucial problems of existing works for TTS that utilize dependency relations based on graph networks.
To address these issues, we proposed Relation-aware Word Encoding Network for text-to-speech synthesis.
RWEN effectively allows linguistic features to utilize dependency relations and can be easily incorporated into most existing TTS models.
Moreover, experimental results show that RWEN outperforms existing works, and we prove that SRE and AWRE are significantly effective through our ablation experiments.
\section{Acknowledgments}
We thank the anonymous reviewers, Kihyuk Jeong, Sunwoo Im for their constructive comments.

\bibliography{aaai23}

\end{document}